\documentclass[runningheads]{llncs}
\usepackage[T1]{fontenc}

\usepackage{amsmath,verbatim,amssymb,amsfonts,amscd,graphicx}
\usepackage[utf8]{inputenc}
\usepackage{longtable}
\usepackage{pgfplots}
\usepackage{multirow}
\usepackage{makecell}
\usepackage{enumitem}
\usepackage{xfrac}
\usepackage[title]{appendix}
\usepackage[english]{babel}
\usepackage{pdflscape}
\usepackage{hyperref}

\usepackage{tikz}
\usetikzlibrary{arrows,shapes,positioning,shadows,trees,calc}
\tikzset{
    basic/.style  = {draw, text width=2cm, drop shadow, font=\sffamily, rectangle},
    root/.style   = {basic, rounded corners=2pt, thin, align=center,
                     fill=green!30},
    level 2/.style = {basic, rounded corners=6pt, thin,align=center, fill=green!60,
                     text width=8em},
    level 3/.style = {basic, thin, align=center, fill=pink!60, text width=9.5em}
}

\usepackage[linesnumbered,ruled]{algorithm2e}

\newcommand{\eps}{\varepsilon}

\newlength{\summtableheight}  
\usepackage{array}
\pgfplotsset{compat=1.17}

\begin{document}

\title{Superior Parallel Big Data Clustering through Competitive Stochastic Sample Size Optimization in Big-means}

\author{Rustam Mussabayev\inst{1,2}\orcidID{0000-0001-7283-5144} \and
Ravil Mussabayev\inst{1,3}\orcidID{0000-0003-1105-5990}}

\authorrunning{R. Mussabayev and R. Mussabayev}

\institute{Satbayev University, Satbayev str. 22, Almaty 050013, Kazakhstan \and Laboratory for Analysis and Modeling of Information Processes, Institute of Information and Computational Technologies, Pushkin str. 125, Almaty 050010, Kazakhstan\\
\email{rustam@iict.kz} \and
University of Washington, Department of Mathematics, Padelford Hall C-138, Seattle 98195-4350, WA, USA\\
\email{ravmus@uw.edu}}

\maketitle

\begin{abstract}
This paper introduces a novel K-means clustering algorithm, an advancement on the conventional Big-means methodology. The proposed method efficiently integrates parallel processing, stochastic sampling, and competitive optimization to create a scalable variant designed for big data applications. It addresses scalability and computation time challenges typically faced with traditional techniques. The algorithm adjusts sample sizes dynamically for each worker during execution, optimizing performance. Data from these sample sizes are continually analyzed, facilitating the identification of the most efficient configuration. By incorporating a competitive element among workers using different sample sizes, efficiency within the Big-means algorithm is further stimulated. In essence, the algorithm balances computational time and clustering quality by employing a stochastic, competitive sampling strategy in a parallel computing setting.

\keywords{Big-means Clustering \and Parallel Computing \and Data Mining \and Stochastic Variation \and Sample Size \and Competitive Environment \and Parallelization Strategy \and Machine Learning \and Big Data Analysis \and Optimization \and Cluster Analysis \and K-means \and K-means++ \and Unsupervised Learning.}
\end{abstract}

\section{Introduction}

Clustering is a foundational task in the field of data analysis and machine learning, serving as the cornerstone of unsupervised learning techniques~\cite{Hastie2009}. The ultimate goal of clustering is to partition a set of objects into groups, known as clusters, in such a way that objects belonging to the same cluster exhibit higher similarity to each other than to those in different clusters~\cite{Jain2010-data}. This task is driven by the inherent human instinct to understand and organize large volumes of information, and it mirrors our natural desire to categorize or group similar objects together.

The importance and ubiquity of clustering in various domains cannot be overstated. For instance, in the realm of image processing and computer vision, clustering algorithms are routinely employed for image segmentation, where similar pixels are grouped together to identify regions of interest~\cite{Comaniciu2002-mean}. In the world of business and marketing, customer segmentation is crucial for targeted marketing, and it heavily relies on clustering methods to identify groups of customers with similar buying patterns or behaviors~\cite{Ng2002-clustering}.

In bioinformatics, clustering is used in gene expression data analysis where genes with similar expression patterns are grouped, leading to the identification of functionally related groups of genes~\cite{Eisen1998-cluster}. In the field of natural language processing, clustering is used to group similar texts, helping in tasks like topic modeling and sentiment analysis~\cite{Deerwester1990-indexing}. Additionally, social network analysis often uses clustering to identify communities or groups of users with similar interests or behaviors~\cite{Fortunato2010-community}.

Despite the diverse applications, all clustering methods share a common goal --- to maximize the intra-cluster similarity and minimize the inter-cluster similarity. Among various clustering methods, one common and widely accepted criterion for clustering is the minimum-sum-of-squares clustering (MSSC) criterion. The MSSC criterion is also known as the K-means criterion, as the popular K-means algorithm optimizes this criterion~\cite{Macqueen1967-some}.

The MSSC criterion aims to partition the data into K clusters, such that the sum of squared Euclidean distances between the data points and their associated cluster centers is minimized~\cite{Forgy1965-cluster}. In mathematical terms, if we denote the center of a cluster by $c \in C$ and the points in a cluster by $x \in X$, then we aim to minimize

\begin{equation}
\min\limits_{C} \ \ \ f\left(C,X\right)=\sum\limits_{i=1}^m \min_{j=1,\ldots,k} \| x_i - c_j \|^2
\label{eq:mssc}
\end{equation}
where $k$ is the number of clusters, and $\| \cdot \|$ stands for the Euclidean norm.

In practice, the MSSC criterion often leads to high-quality clusters when the underlying clusters in the data are spherical and evenly sized~\cite{Ding2004-k}. However, the criterion and its associated algorithms are sensitive to the initial configuration and are susceptible to getting trapped in local minima~\cite{Arthur2007}. This motivates the search for advanced and robust clustering algorithms that can overcome these limitations~\cite{Celebi2013-comparative}.

The purpose of this paper is to introduce a novel parallel clustering algorithm with competitive stochastic sample size optimization and evaluate its performance. Competitive optimization involves multiple solutions or agents (parallel processes or algorithms) competing to achieve optimal results, leading to a dynamic and adaptive optimization process. Stochastic sampling, the random selection of data subsets, is used by each process to reduce computational load while maintaining a representative data set. The new algorithm is a parallel version of the Big-means clustering algorithm~\cite{Mussabayev2023,Mussabayev2023-parbigmeans}, enhanced with competitive stochastic sample size optimization. Each parallel process competes using different sample sizes for K-means clustering, striving for the most efficient clustering outcome. Results of extensive experiments suggest that this approach is more efficient and scales better with large datasets compared to traditional K-means and all of its variants.

The rest of the paper is structured as follows. Section~\ref{sec:problem_statement} motivates the creation of the new algorithm. Section~\ref{sec:related_works} provides a comprehensive literature review of available methods, highlighting their main shortcomings. Section~\ref{sec:methodology} presents the proposed method in detail. Section~\ref{sec:experiments} describes the experimental setup used to evaluate the performance of the algorithm. Section~\ref{sec:exp_results} discusses the results and compares the new algorithm with existing methods. Finally, Section~\ref{sec:conclusion} concludes the paper and suggests directions for future work.

\section{Problem Statement} \label{sec:problem_statement}

As the volume of data being generated and collected continues to grow exponentially, the demand for efficient, scalable, and robust algorithms for data analysis tasks like clustering has been significantly heightened. This is especially true for MSSC  algorithms that can handle large datasets, often referred to as big data. Despite the ubiquity of MSSC clustering in various domains, several challenges hinder the efficient execution of these algorithms on big datasets. 

Firstly, many of the existing MSSC algorithms do not scale well with the size of the dataset. The computational complexity of these algorithms grows rapidly as the number of data points increases, rendering them impractical for use on big data. Notably, the popular K-means algorithm, while simple and effective on smaller datasets, suffers from a high computational cost when dealing with large-scale data due to the need to compute distances between each data point and each centroid in each iteration.

Secondly, several current algorithms do not fully harness the potential of modern high-performance computing systems. With the advancements in multi-core and multi-processor architectures, as well as the advent of distributed computing platforms, there exists a great opportunity to develop parallel algorithms for MSSC that can process large datasets efficiently by utilizing all available computational resources. However, the development of such parallel algorithms is not straightforward and involves tackling challenges like data partitioning, load balancing, and communication overhead among the processing units.

Finally, a persistent challenge in MSSC is the sensitivity of the algorithms to the initial configuration, which can lead to sub-optimal clustering solutions trapped in local minima. This problem becomes more pronounced in the case of big data due to the increased complexity and diversity of the data.

Therefore, the need for a new MSSC clustering algorithm that is both efficient and scalable, can handle big data, takes full advantage of modern high-performance computing systems, and provides robust clustering solutions regardless of the initial configuration, is evident. Such an algorithm would significantly enhance our ability to extract valuable insights from large datasets in various fields.



\section{Related Works} \label{sec:related_works}

Clustering has been an intensely studied area in machine learning, and numerous algorithms have been proposed to solve the problem of minimum-sum-of-squares clustering. The K-means algorithm, owing to its simplicity, is the most popular and widely used of these algorithms~\cite{Macqueen1967-kmeans}. However, K-means has several notable limitations. It tends to get stuck in local minima, is sensitive to the initial placement of centroids, and its performance degrades with an increase in the size of the dataset.

To overcome these limitations, several variants of K-means have been proposed. The K-means++ algorithm introduces a smarter initialization technique that aims to provide better initial centroids, thereby improving the final clustering result~\cite{Arthur2007}. While this technique reduces the risk of getting stuck in poor local minima, it does not adequately address scalability issues.

MiniBatch K-means is a variant designed to handle larger datasets more efficiently by operating on a randomly selected subset of the data at each step~\cite{Sculley2010-minibatch}. While this improves computational efficiency, it comes at the cost of potentially reduced clustering quality compared to standard K-means.

Scalable K-means++, as the name suggests, is designed to scale better with big data. This variant selects initial centers in an ``oversampling'' phase and then reduces the number of centers in a ``reduce phase'', improving scalability. However, the complexity of the algorithm and the need for multiple passes over the data limit its practicality~\cite{Bahmani2012-scalablekmeans}.

Bisecting K-means adopts a divide-and-conquer strategy where the algorithm iteratively splits clusters~\cite{Steinbach2000-bisecting}. This can potentially result in a better quality of clustering but requires more computational resources as the number of clusters or the size of the data increases.

Fuzzy C-means provides a probabilistic approach where each data point can belong to multiple clusters with varying membership grades~\cite{Bezdek1981-pattern}. This can offer more nuanced cluster assignments but increases the computational complexity significantly.

X-means and G-means algorithms try to determine the appropriate number of clusters automatically, thereby removing the need for explicit cluster number specification~\cite{Pelleg2000-xgmeans}. However, this automation introduces its complexities and can result in increased computational cost.

Hierarchical K-means presents a hierarchical approach to clustering, which can handle larger datasets and provide a more intuitive cluster hierarchy~\cite{Zhang1996-hkm}. Yet, it suffers from high computational costs for large datasets.

Finally, Distributed K-means algorithms take advantage of distributed computing resources to scale the K-means clustering process to larger datasets~\cite{Zaharia2012-dkm}. While this approach effectively utilizes modern computational resources, the communication overhead and the need for data to be partitioned appropriately can limit its efficiency.

Despite the strengths of these algorithms, they often introduce additional computational complexity or do not adequately scale with larger datasets. Hence, the need for a new algorithm that efficiently handles large datasets while taking full advantage of modern high-performance computing resources is evident.

\section{Methodology} \label{sec:methodology}

\subsection{Big-means Algorithm}

The Big-means algorithm is specifically designed to tackle large-scale Minimum Sum-of-Squares Clustering (MSSC) problems. The concept underlying the Big-means algorithm is straightforward~\cite{Mussabayev2023}: in each iteration, a new uniformly random sample of size $s \ll |X|$ is drawn from the provided dataset and clustered using K-means. The K-means++ algorithm initializes clustering for the first sample. Every subsequent sample is initialized using the best solution found thus far across previous iterations in terms of minimizing the objective function~\eqref{eq:mssc} on the samples. During intermediate iterations, only degenerate clusters are reinitialized using K-means++. Iterations persist until a ``stop condition'' is met. This ``stop condition'' can be a limit on CPU time or a maximum number of samples to be processed. The outcome of the algorithm is the set of centroids that achieved the best objective function value~\eqref{eq:mssc} across the iterations. Ultimately, all data points can be allocated to clusters based on their proximity to the resulting centroids.

The shaking procedure is a vital component of the Big-means algorithm. It refers to the generation of a new sample in each iteration, which perturbs the current (incumbent) solution and introduces variability into the clustering results. When the full dataset is viewed as a cloud of points in an $n$-dimensional space, each sample represents only a sparse approximation of this cloud. This procedure instills diversity and adaptability into the clustering process. It accomplishes this by iteratively applying the K-means algorithm to random subsets of the data, progressively refining the centroid locations, and effectively managing degenerate clusters.

The algorithm's scalability can be fine-tuned by selecting appropriate sample sizes and counts. Processing smaller subsets of data in each iteration significantly reduces computational demands. Additionally, this strategy prevents the algorithm from getting trapped in suboptimal solutions. This is achieved by using random data subsets in each iteration and periodically re-initializing the centroids of degenerate clusters. The Big-means algorithm is a promising approach for clustering large datasets, offering scalability, efficiency, and robustness.


\subsection{Competitive Sample Size Big-means Algorithm}

\begin{algorithm}[!htbp]
\SetAlgoLined
\textbf{Initialization:}\\
$C_w \leftarrow \text{Mark all } k \text{ centroids as degenerate for each worker } w$\;
$\hat{f}_w \leftarrow \infty$ for each worker $w$\;
$t_w \leftarrow 0$ for each worker $w$\;
$L \leftarrow \text{Empty list}$\;
\While{$t_w < T$ for any worker $w$}{
    \For{each parallel worker $w$}{
        $s_{w} \leftarrow \text{Random integer in } [s_{min}, s_{max}]$\;
        Recalculate $\hat{f}_w$ with the new $s_{w}$\;
        $p_w \leftarrow 0$\;
        \While{$p_w < p$}{
            $S_{w} \leftarrow \text{Random sample of size } s_{w} \text{ from } X$\;
            \For{each $c \in C_w$}{
                \If{$c$ is the centroid associated with a degenerate cluster}{
                    \text{Reinitialize }$c \text{ using K-means++ on } S_{w}$\;
                }
            }
            $C_{\text{new},w} \leftarrow \text{K-means clustering on } S_{w} \text{ with initial centroids } C_w$\;
            \If{$f(C_{\text{new},w}, S_{w}) < \hat{f}_w$}{
                $C_w \leftarrow C_{\text{new},w}$\;
                $\hat{f}_w \leftarrow f(C_{\text{new},w}, S_{w})$\;
                Add $s_{w}$ to list $L$\;
            }
            $p_w \leftarrow p_w + 1$\;
        }
        $t_w \leftarrow t_w + 1$\;
    }
}
Analyze distribution of $s_i$ values in list $L$\;
$s_{opt} \leftarrow s_i$ value with highest probability of improving objective function\;
$S \leftarrow \text{Random sample of size } s_{opt} \text{ from } X$\;

\For{each parallel worker $w$}{
    Recalculate $\hat{f}_w$ with $s_{opt}$ using $S$\;
}

$C_{\text{best}} \leftarrow \text{Centroids of the worker with the smallest } \hat{f}_w \text{ value}$\;
$Y \leftarrow \text{Assign each point in } X \text{ to nearest centroid in } C_{\text{best}}$\;
\Return{$C_{\text{best}}$, $Y$, $s_{opt}$}\;
\caption{Detailed pseudocode of Parallel Big-means Clustering with Competitive Stochastic Sample Size Optimization}
\label{alg:competitive_sample_size_big_means_detailed}
\end{algorithm}


In our study, we propose a novel strategy to parallelize the Big-means clustering algorithm, wherein the size $s$ of the clustered sample from big data used at each iteration varies. Within this parallelization strategy, during the algorithm's initialization stage, each of the $w$ workers randomly selects their sample size $s_{w}$ from the permissible range $[s_{min}; s_{max}]$. Subsequently, each worker operates in parallel for $p$ iterations, adhering to the standard Big-means algorithm's scheme using the allocated sample size $s_{w}$.

After $p$ iterations, each $i$-th worker is assigned a new random sample size $s_{w}$. Simultaneously, the value of the target criterion for the worker's current incumbent solution is recalculated to reflect the change in $s_{w}$, as any change in $s_{w}$ necessitates such a recalculation. As the algorithm operates, comprehensive statistics relating to improvements in the objective function's value for a given $s_{w}$ are collected for all workers. Essentially, each worker, upon witnessing an improvement in the value of its objective function, contributes its current $s_{w}$ value to a shared list.

By the end of the algorithm's execution, we obtain a list of all $s_{w}$ values that led to improvements in the objective function's value. By analyzing the distribution of values in this list, we select the $s_{opt}$ value from the list that offers the highest probability of enhancing the objective function's value. The desired result can be achieved by calculating a simple mean of the given list, which corresponds to the expected value of the improving sample size.

This strategy effectively creates a competitive environment among workers, allowing for simultaneous variation in the used sample size $s_{w}$ and subsequent determination of its optimal value.

\subsection{Detailed Algorithm Description}

The algorithm presented in Algorithm~\ref{alg:competitive_sample_size_big_means_detailed} details a parallel implementation of the K-means clustering method with competitive stochastic sample size optimization. The method aims to determine the best cluster centroids and assign data points to these clusters efficiently.

The algorithm initializes by marking all $k$ centroids as degenerate for each worker $w$. Each worker also has an initial best-so-far objective function value $\hat{f}_w$ and iteration count $t_w$ set to $\infty$ and 0, respectively. An empty list $L$ is also defined to keep track of sample sizes that lead to improvement in the objective function.

In each iteration of the main loop, every worker operates in parallel, choosing a random sample size $s_{w}$ between $s_{min}$ and $s_{max}$, and recomputing the $\hat{f}_w$ with the new sample size. Then, within the defined maximum number of passes $p$, the worker takes a random sample of size $s_{w}$ from the data set $X$ and processes the centroids. For each centroid $c$ associated with a degenerate cluster, the worker reinitializes $c$ using K-means++ on the sample $S_{w}$. The worker performs K-means clustering on $S_{w}$ with initial centroids $C_w$ to get new centroids $C_{\text{new},w}$. If the new centroids result in a better objective function value, the worker updates its current centroids, best-so-far objective function value, and adds the sample size to list $L$. The pass counter $p_w$ is then incremented.

After all workers finish their iterations or reach the maximum number of iterations $T$, the algorithm proceeds to analyze the distribution of sample sizes in list $L$. The algorithm selects the sample size $s_{opt}$, which has the highest probability of improving the objective function. This selection can be achieved by calculating the simple mean of the values in the list $L$. Then, a new sample of size $s_{opt}$ is taken from the dataset $X$, and each worker recalculates its best-so-far objective function value with this new sample size.

Finally, the algorithm chooses the centroids $C_{\text{best}}$ of the worker with the smallest $\hat{f}_w$ value, assigns each point in $X$ to its nearest centroid in $C_{\text{best}}$, and returns these centroids, the cluster assignments, and the optimal sample size $s_{opt}$.

In this article, we assume that each worker has equal access to the full-sized dataset and can independently draw samples from it. For the sake of simplicity, in this study we are not exploring various available opportunities for further optimization of the algorithm, particularly those concerning distributed data storage across different nodes of the computing system. Such optimizations merit a separate study.

\section{Experiment Setup} \label{sec:experiments}

Our experiments utilized a system with Ubuntu 22.04 (64-bit), an AMD EPYC 7663 56-Core Processor, and 1.46 TB RAM. Up to 16 cores were deployed, running Python 3.10.11, NumPy 1.24.3, and Numba 0.57.0. Numba~\cite{Marowka2018} was essential for Python code acceleration and parallelism.

The study involved comparing the proposed algorithm to the best-performing hybrid-parallel version of the Big-means algorithm~\cite{Mussabayev2023-parbigmeans}, using 19 public datasets (details and URLs are available in~\cite{Mussabayev2023} and \cite{Mussabayev2023-parbigmeans}), plus four normalized datasets, totaling 23. These datasets, with 2 to 5,000 attributes and 7,797 to 10,500,000 instances, were used to assess our algorithm's versatility.

We executed each algorithm on the 23 datasets $n_{exec}$ times for cluster sizes of 2, 3, 5, 10, 15, 20, 25. Select datasets underwent additional clustering into 4 clusters, following Karmitsa et al.~\cite{Karmitsa2018}. This resulted in 7,366 individual clustering processes. The algorithms' performances were measured in terms of relative clustering accuracy $\eps$, CPU time $t$, and baseline time $\overline{t}$.

The relative clustering accuracy $\eps$ is defined as
$$
\eps(\%) = \frac{100 \times (f - f^*)}{f^*},
$$
with $f^* = f^*(X, k)$ being the best historical objective function value for dataset $X$ and cluster number $k$.

To objectively measure clustering time, we use a special baseline time metric $\overline{t}$, which helps to avoid bias from minor late-stage improvements. For every algorithm, time $\overline{t}$ is the time it takes for an algorithm to achieve the baseline sample objective value $\overline{f}_s$. For every pair $(X, k)$ and sample size $s$, $\overline{f}_s$ is derived from comparing the sample objective values achieved by different Big-means parallelization strategies: sequential, inner, competitive, and collective~\cite{Mussabayev2023-parbigmeans}. Specifically, $\overline{f}_s$ is defined as the maximum of the median (median is taken over $n_{exec}$ executions for the pair $(X, k)$) best sample objective function values across iterations obtained by the considered parallel Big-means versions. Essentially, this method determines the worst-performing parallel Big-means version and treats its accuracy on the best processed sample as the baseline. In multi-worker settings, $\overline{t}$ is the time taken by the fastest worker to reach $\overline{f}_s$.

Clustering was limited to 300 iterations or a relative tolerance of $10^{-4}$, and K-means++ was used with three candidate points for centroid generation. For the proposed algorithm, $s_{min}$ was chosen to be $0.5$ of the sample size $s$ used in Big-means, while $s_{max}$ was chosen to be 2 times $s$ (or $m$, if this number exceeds $m$ for the given dataset). We set the parameter value $p = 10$ for all experiments. For each pair $(X, k)$, the choice of parameters $s, t_{max}$ and $n_{exec}$ precisely matched the values specified in the original Big-means paper~\cite{Mussabayev2023}.

Preliminary experiments established the optimal number of CPUs, baselines, as well as the optimal parameter values for the hybrid parallel version of Big-means, as described in~\cite{Mussabayev2023-parbigmeans}. The main experiment used these baselines and parameter values. We calculated the minimum, median, and maximum values of relative accuracy and CPU time for each $(X, k)$ pair over $n_{exec}$ runs. The results are summarized in Tables~\ref{tab:result_e} -- \ref{tab:result_t_bar}. These tables highlight top performances, providing a comprehensive evaluation across all datasets.

\section{Experimental Results and Discussion} \label{sec:exp_results}

\subsection{Performance Evaluation}

\begin{table}[!htbp]%
\centering%
\caption{Relative clustering accuracies $\epsilon$ (\%) for different algorithms. The highest accuracies for each experiment (algorithm, data pair $(X, k)$) are displayed in bold. Success is indicated when an algorithm's performance matches the best result among all algorithms for the current experiment.}%
\label{tab:result_e}%
\resizebox{!}{\summtableheight}{%
\begin{tabular}{l|cccc|cccc}
\hline
\multirow{2}{*}{Dataset} & \multicolumn{4}{p{3cm}}{\mbox{Proposed algorithm}}& \multicolumn{4}{p{3cm}}{\mbox{Big-means}} \\
\cline{2-9}
& \#Succ & Min & Median & Max & \#Succ & Min & Median & Max  \\
\hline
CORD-19 Embeddings & 45/49 & \textbf{-0.07} & \textbf{0.01} & \textbf{0.21} & 4/49 & 0.01 & 0.03 & 0.38 \\
HEPMASS & 41/49 & \textbf{-0.07} & \textbf{0.04} & \textbf{0.33} & 8/49 & 0.0 & 0.12 & 0.44 \\
US Census Data 1990 & 102/140 & \textbf{0.0} & \textbf{1.43} & \textbf{4.74} & 38/140 & 0.03 & 1.97 & 5.6 \\
Gisette & 97/105 & \textbf{-1.86} & \textbf{0.0} & \textbf{0.1} & 8/105 & -1.71 & 0.01 & 0.29 \\
Music Analysis & 135/140 & \textbf{0.01} & \textbf{0.23} & \textbf{1.3} & 5/140 & 0.02 & 0.54 & 3.57 \\
Protein Homology & 103/105 & \textbf{-0.05} & \textbf{0.05} & \textbf{1.83} & 2/105 & 0.11 & 0.8 & 2.69 \\
MiniBooNE Particle Identification & 92/105 & \textbf{-0.54} & \textbf{0.0} & \textbf{2.36} & 13/105 & -0.38 & 0.01 & \textbf{2.36} \\
MiniBooNE Particle Identification (normalized) & 139/140 & \textbf{-0.0} & \textbf{0.14} & \textbf{0.8} & 1/140 & 0.01 & 0.41 & 3.16 \\
MFCCs for Speech Emotion Recognition & 131/140 & \textbf{0.0} & \textbf{0.05} & \textbf{1.99} & 9/140 & 0.02 & 0.11 & 2.21 \\
ISOLET & 86/105 & \textbf{-0.49} & \textbf{0.01} & \textbf{0.67} & 19/105 & -0.15 & 0.25 & 1.66 \\
Sensorless Drive Diagnosis & 224/280 & \textbf{-2.42} & \textbf{-0.0} & \textbf{5.58} & 56/280 & -2.41 & \textbf{-0.0} & 162.06 \\
Sensorless Drive Diagnosis (normalized) & 237/280 & \textbf{0.0} & \textbf{1.02} & \textbf{4.45} & 43/280 & 0.01 & 1.28 & 7.87 \\
Online News Popularity & 130/140 & \textbf{-0.39} & \textbf{0.17} & \textbf{5.02} & 10/140 & 0.01 & 0.88 & 11.59 \\
Gas Sensor Array Drift & 166/210 & \textbf{-0.92} & \textbf{0.04} & \textbf{4.06} & 44/210 & -0.77 & 0.26 & 8.42 \\
3D Road Network & 267/280 & \textbf{0.0} & \textbf{0.04} & \textbf{0.66} & 13/280 & \textbf{0.0} & 0.22 & 2.76 \\
Skin Segmentation & 181/210 & \textbf{-1.38} & \textbf{0.1} & \textbf{5.34} & 29/210 & -1.3 & 0.21 & 10.01 \\
KEGG Metabolic Relation Network (Directed) & 114/140 & \textbf{-1.27} & \textbf{0.0} & \textbf{2.93} & 26/140 & -1.24 & 0.03 & 27.4 \\
Shuttle Control & 99/120 & \textbf{-3.22} & \textbf{0.0} & \textbf{6.93} & 21/120 & -3.12 & 1.38 & 16.48 \\
Shuttle Control (normalized) & 145/160 & \textbf{0.01} & \textbf{0.23} & \textbf{5.59} & 15/160 & 0.05 & 1.34 & 10.43 \\
EEG Eye State & 143/160 & \textbf{-0.09} & \textbf{0.0} & \textbf{4.25} & 17/160 & -0.06 & 0.01 & 29.91 \\
EEG Eye State (normalized) & 212/240 & \textbf{-0.38} & \textbf{-0.0} & \textbf{0.7} & 28/240 & -0.33 & \textbf{0.0} & 598.89 \\
Pla85900 & 264/280 & \textbf{-0.06} & \textbf{0.04} & \textbf{1.44} & 16/280 & 0.0 & 0.12 & 1.6 \\
D15112 & 100/105 & \textbf{-0.02} & \textbf{0.01} & \textbf{0.69} & 5/105 & 0.01 & 0.12 & 1.33 \\
\hline
Overall Results & \textbf{3253/3683} & \textbf{-0.57} & \textbf{0.29} & \textbf{2.69} & 430/3683 & -0.49 & 0.66 & 39.61 \\ \hline
\end{tabular}%
}

\bigskip

\caption{Resulting clustering times $\overline{t}$ (sec.) with respect to baseline sample objective values $\overline{f}_s$. The lowest clustering times for each experiment (algorithm, data pair $(X, k)$) are displayed in bold.}%
\label{tab:result_t_bar}%
\resizebox{!}{\summtableheight}{%
\begin{tabular}{l|ccc|ccc}
\hline
\multirow{2}{*}{Dataset} & \multicolumn{3}{p{3cm}}{\mbox{Proposed algorithm}}& \multicolumn{3}{p{3cm}}{\mbox{Big-means}} \\
\cline{2-7}
& Min & Median & Max & Min & Median & Max  \\
\hline
CORD-19 Embeddings & \textbf{1.69} & 13.2 & 41.17 & 2.14 & \textbf{13.19} & \textbf{40.1} \\
HEPMASS & 1.5 & 4.77 & 24.6 & \textbf{0.75} & \textbf{2.77} & \textbf{21.53} \\
US Census Data 1990 & 0.46 & 2.07 & 5.21 & \textbf{0.09} & \textbf{0.74} & \textbf{2.75} \\
Gisette & \textbf{2.38} & \textbf{14.53} & \textbf{43.52} & 5.07 & 26.14 & 56.92 \\
Music Analysis & \textbf{0.19} & \textbf{3.49} & 14.33 & 0.6 & 3.58 & \textbf{12.87} \\
Protein Homology & \textbf{0.08} & \textbf{3.24} & 13.25 & 0.37 & 3.69 & \textbf{10.43} \\
MiniBooNE Particle Identification & \textbf{0.59} & 4.92 & 15.56 & 0.62 & \textbf{4.71} & \textbf{15.48} \\
MiniBooNE Particle Identification (normalized) & \textbf{0.02} & \textbf{0.6} & \textbf{1.82} & 0.03 & 0.74 & 2.04 \\
MFCCs for Speech Emotion Recognition & \textbf{0.03} & \textbf{0.67} & 2.34 & 0.23 & 0.81 & \textbf{1.62} \\
ISOLET & \textbf{0.04} & \textbf{1.81} & 7.29 & 0.62 & 2.57 & \textbf{6.96} \\
Sensorless Drive Diagnosis & 0.22 & 1.72 & 4.86 & \textbf{0.08} & \textbf{1.63} & \textbf{4.62} \\
Sensorless Drive Diagnosis (normalized) & \textbf{0.01} & \textbf{0.21} & 0.9 & \textbf{0.01} & 0.28 & \textbf{0.56} \\
Online News Popularity & \textbf{0.01} & 0.53 & \textbf{1.47} & 0.03 & \textbf{0.52} & 1.56 \\
Gas Sensor Array Drift & \textbf{0.01} & \textbf{0.7} & \textbf{2.23} & 0.22 & 1.15 & 3.13 \\
3D Road Network & \textbf{0.02} & \textbf{0.63} & \textbf{2.2} & 0.03 & 0.69 & 2.52 \\
Skin Segmentation & \textbf{0.01} & \textbf{0.06} & 0.61 & \textbf{0.01} & 0.17 & \textbf{0.39} \\
KEGG Metabolic Relation Network (Directed) & \textbf{0.02} & 0.77 & \textbf{1.96} & \textbf{0.02} & \textbf{0.69} & 2.11 \\
Shuttle Control & \textbf{0.03} & \textbf{0.54} & 1.56 & 0.21 & 0.62 & \textbf{1.29} \\
Shuttle Control (normalized) & \textbf{0.0} & \textbf{0.05} & \textbf{0.39} & 0.01 & 0.27 & 0.4 \\
EEG Eye State & 0.07 & \textbf{0.48} & 1.35 & \textbf{0.04} & 0.58 & \textbf{1.2} \\
EEG Eye State (normalized) & \textbf{0.01} & 0.18 & 0.84 & \textbf{0.01} & \textbf{0.16} & \textbf{0.77} \\
Pla85900 & \textbf{0.01} & \textbf{0.17} & \textbf{0.68} & \textbf{0.01} & 0.57 & 1.51 \\
D15112 & \textbf{0.01} & \textbf{0.04} & \textbf{0.39} & 0.03 & 0.38 & 1.04 \\
\hline
Overall Results & \textbf{0.32} & \textbf{2.64} & \textbf{8.2} & 0.49 & 3.04 & 8.34 \\ \hline
\end{tabular}%
}
\end{table}

A summary of the results of the main experiment are provided in Tables~\ref{tab:result_e} -- \ref{tab:result_t_bar}. Based on the experimental results, it was observed that the proposed algorithm performed consistently better than Big-means on all datasets, both with respect to the accuracy and time.


We attribute the outstanding performance of the proposed algorithm to its ability to approximate the probability distribution of sample sizes that improve the sample objective function. In the competitive parallelization strategy, each worker starts with its own K-means++ initialization, which strengthens the final result via diversification. Also, due to this kind of parallelism, the algorithm is able to accumulate the necessary statistics for the approximation in a very timely efficient manner. Then, the optimal sample size value $s_{opt}$ can be obtained by evaluating the simple mean of the accumulated sample size occurrence distribution.

The value $s_{opt}$ is the size of a sample that attains the best balance between sparsifying high-density clusters while still including enough mass of low-density clusters into the sample. In addition to the accumulation of the improving sample size distribution, competitive workers are able to dynamically guide the flow of centroids through unfavorable situations by using various random sample sizes in the range $[s_{min}, s_{max}]$. For instance, these unfavorable situations might occur when a sample has large gaps between highly dense clusters (thus preventing the fluidity of centroids between them) or excludes some clusters due to an overly intense sparsification.

\section{Conclusion and Future Works} \label{sec:conclusion}

In this work, we proposed a parallel Big-means clustering algorithm equipped with competitive stochastic sample size optimization, as well as thoroughly evaluated its performance against the state-of-the-art hybrid-parallel version of the Big-means algorithm~\cite{Mussabayev2023-parbigmeans} using a wide array of real-world datasets from the Big-means' original paper~\cite{Mussabayev2023}. The idea of using an automatic procedure for approximating the optimal sample size stemmed from multiple considerations. First, it is practically hard and error-prone to estimate the sample size for real-world big datasets. Second, using a fixed sample size makes the iterative improvement nature of the Big-means algorithm too rigid. Indeed, sampling with a fixed sample size is limited in the flexibility of approximating and sparsifying regions of the dataset with different densities.

Our improved version of Big-means exhibited exceptional results both in the resulting quality and time, pushing much further the state of the art in the field of big data clustering. We are confident that our work presents a valuable contribution to the scientific field, as well as brings a tool of considerable practical value to practitioners in the field of big data.

For future research, we plan to investigate other dimensions for experimentation, including different ways to explore the range $[s_{min}, s_{max}]$ and exploit the currently best obtained sample size across iterations.

\section*{Acknowledgements}

This research was funded by the Science Committee of the Ministry of Science and Higher Education of the Republic of Kazakhstan (grant no. BR21882268).

\bibliography{main}

\begin{thebibliography}{10}
\providecommand{\url}[1]{\texttt{#1}}
\providecommand{\urlprefix}{URL }
\providecommand{\doi}[1]{https://doi.org/#1}

\bibitem{Arthur2007}
Arthur, D., Vassilvitskii, S.: K-means++: The advantages of careful seeding.
  In: Proceedings of the Eighteenth Annual ACM-SIAM Symposium on Discrete
  Algorithms. p. 1027–1035. SODA '07, Society for Industrial and Applied
  Mathematics, USA (2007)

\bibitem{Bahmani2012-scalablekmeans}
Bahmani, B., Moseley, B., Vattani, A., Kumar, R., Vassilvitskii, S.: Scalable
  k-means++. Proceedings of the VLDB Endowment  \textbf{5}(7),  622--633 (2012)

\bibitem{Bezdek1981-pattern}
Bezdek, J.C.: Pattern Recognition with Fuzzy Objective Function Algorithms.
  Plenum Press, New York (1981)

\bibitem{Celebi2013-comparative}
Celebi, M.E.: Comparative Performance of Seeding Methods for k-Means Algorithm.
  Springer (2013)

\bibitem{Comaniciu2002-mean}
Comaniciu, D., Meer, P.: Mean shift: a robust approach toward feature space
  analysis. IEEE Transactions on Pattern Analysis and Machine Intelligence
  \textbf{24}(5),  603--619 (2002)

\bibitem{Deerwester1990-indexing}
Deerwester, S., Dumais, S.T., Furnas, G.W., Landauer, T.K., Harshman, R.:
  Indexing by latent semantic analysis. Journal of the American Society for
  Information Science  \textbf{41}(6) (1990)

\bibitem{Ding2004-k}
Ding, C., He, X.: K-means clustering via principal component analysis. In:
  Proceedings of the twenty-first international conference on Machine learning
  (2004)

\bibitem{Eisen1998-cluster}
Eisen, M.B., Spellman, P.T., Brown, P.O., Botstein, D.: Cluster analysis and
  display of genome-wide expression patterns. Proceedings of the National
  Academy of Sciences  \textbf{95}(25),  14863--14868 (1998)

\bibitem{Forgy1965-cluster}
Forgy, E.W.: Cluster analysis of multivariate data: efficiency vs
  interpretability of classifications. Tech. Rep. RM-5437-PR, RAND Corporation
  (1965)

\bibitem{Fortunato2010-community}
Fortunato, S.: Community detection in graphs. Physics Reports
  \textbf{486}(3-5),  75--174 (2010)

\bibitem{Hastie2009}
Hastie, T., Tibshirani, R., Friedman, J.: The elements of statistical learning:
  data mining, inference, and prediction. Springer Science \& Business Media
  (2009)

\bibitem{Jain2010-data}
Jain, A.K.: Data clustering: 50 years beyond k-means. Pattern Recognition
  Letters  \textbf{31},  651--666 (2010)

\bibitem{Karmitsa2018}
Karmitsa, N., Bagirov, A.M., Taheri, S.: Clustering in large data sets with the
  limited memory bundle method. Pattern Recognition  (2018)

\bibitem{Macqueen1967-some}
MacQueen, J.: Some methods for classification and analysis of multivariate
  observations. In: Proceedings of the fifth Berkeley symposium on mathematical
  statistics and probability. vol.~1, pp. 281--297 (1967)

\bibitem{Macqueen1967-kmeans}
MacQueen, J.B.: Some methods for classification and analysis of multivariate
  observations  \textbf{1}(14) (1967)

\bibitem{Marowka2018}
Marowka, A.: Python accelerators for high-performance computing. The Journal of
  Supercomputing  \textbf{74}(4),  1449--1460 (2018)

\bibitem{Mussabayev2023-parbigmeans}
Mussabayev, R., Mussabayev, R.: Strategies for parallelizing the big-means
  algorithm: A comprehensive tutorial for effective big data clustering (2023)

\bibitem{Mussabayev2023}
Mussabayev, R., Mladenovic, N., Jarboui, B., Mussabayev, R.: How to use k-means
  for big data clustering? Pattern Recognition  \textbf{137},  109269 (2023).
  \doi{10.1016/j.patcog.2022.109269}

\bibitem{Ng2002-clustering}
Ng, A.Y., Jordan, M.I., Weiss, Y.: On spectral clustering: Analysis and an
  algorithm. In: Advances in neural information processing systems. pp.
  849--856 (2002)

\bibitem{Pelleg2000-xgmeans}
Pelleg, D., Moore, A.: X-means: Extending k-means with efficient estimation of
  the number of clusters. In Proceedings of the 17th International Conf. on
  Machine Learning pp. 727--734 (2000)

\bibitem{Sculley2010-minibatch}
Sculley, D.: Web-scale k-means clustering. Proceedings of the 19th
  international conference on World wide web pp. 1177--1178 (2010)

\bibitem{Steinbach2000-bisecting}
Steinbach, M., Karypis, G., Kumar, V.: A comparison of document clustering
  techniques. In: KDD Workshop on Text Mining (2000)

\bibitem{Zaharia2012-dkm}
Zaharia, M., Chowdhury, M., Das, T., Dave, A., Ma, J., McCauly, M., Franklin,
  M.J., Shenker, S., Stoica, I.: Resilient distributed datasets: A
  fault-tolerant abstraction for in-memory cluster computing. In: Proceedings
  of the 9th USENIX conference on Networked Systems Design and Implementation.
  USENIX Association (2012)

\bibitem{Zhang1996-hkm}
Zhang, T., Ramakrishnan, R., Livny, M.: Birch: An efficient data clustering
  method for large databases. ACM SIGMOD Record  \textbf{25}(2),  103--114
  (1996)

\end{thebibliography}

\end{document}